\documentclass[dvipsnames,table]{article}
\usepackage[preprint]{colm2025_conference}
\usepackage{microtype}
\usepackage{url}
\usepackage{booktabs}
\usepackage{comment}
\usepackage{times}
\usepackage{latexsym}
\usepackage{graphicx}
\usepackage{colortbl}
\usepackage{multirow}
\usepackage{siunitx}
\usepackage{subcaption}
\usepackage[export]{adjustbox}
\usepackage{dirtytalk}
\usepackage{algorithm}
\usepackage{algpseudocode}
\usepackage{float}
\usepackage{pdflscape}
\usepackage[]{acronym}
\usepackage{amsmath}

\usepackage{amssymb}

\usepackage[T1]{fontenc}
\usepackage[utf8]{inputenc}
\usepackage{inconsolata}
\usepackage{lineno}

\usepackage{hyperref}
\definecolor{darkblue}{rgb}{0, 0, 0.5}
\hypersetup{colorlinks=true, citecolor=darkblue, linkcolor=darkblue, urlcolor=darkblue}

\title{Statistical Multicriteria Evaluation of LLM-Generated Text 
 
}

\author{\textbf{Esteban Garces Arias}$^{1,3}$ \quad
\textbf{Hannah Blocher}$^{1}$ \quad
\textbf{Julian Rodemann}$^{1}$ \\
\textbf{Matthias Aßenmacher}$^{1,3}$ \quad
\textbf{Christoph Jansen}$^{2}$ \\
\\
$^{1}$Department of Statistics, LMU Munich, Germany\\
$^{2}$School of Computing \& Communications, Lancaster University Leipzig, Germany \\
$^{3}$Munich Center for Machine Learning (MCML), Germany \\
\\
\texttt{Esteban.GarcesArias@stat.uni-muenchen.de}
}

\begin{document}

\ifcolmsubmission
\linenumbers
\fi

\maketitle

\begin{abstract}
Assessing the quality of LLM-generated text remains a fundamental challenge in natural language processing. Current evaluation approaches often rely on isolated metrics or simplistic aggregations that fail to capture the nuanced trade-offs between coherence, diversity, fluency, and other relevant indicators of text quality. In this work, we adapt a recently proposed framework for statistical inference based on Generalized Stochastic Dominance (GSD) that addresses three critical limitations in existing benchmarking methodologies: the inadequacy of single-metric evaluation, the incompatibility between cardinal automatic metrics and ordinal human judgments, and the lack of inferential statistical guarantees. The GSD-front approach enables simultaneous evaluation across multiple quality dimensions while respecting their different measurement scales, building upon partial orders of decoding strategies, thus avoiding arbitrary weighting of the involved metrics. By applying this framework to evaluate common decoding strategies against human-generated text, we demonstrate its ability to identify statistically significant performance differences while accounting for potential deviations from the i.i.d. assumption of the sampling design. 
\end{abstract}

\section{Introduction}\label{sec:intro}

Large language models \citep[LLMs; ][]{achiam2023gpt,grattafiori2024llama,guo2025deepseek} have gained widespread popularity due to their remarkable ability to generate coherent and contextually appropriate text across diverse domains. These models fundamentally rely on decoding strategies—algorithmic approaches that determine the selection of each subsequent token based on the model's probability distribution over the vocabulary space. As the architectural complexity and capabilities of these models have advanced, the field has witnessed a proliferation of decoding methodologies, encompassing both deterministic approaches \citep[e.g.][]{Freitag_2017,su2022contrastive} and stochastic variants \citep[e.g.][]{fan2018hierarchical,holtzman2019curious}. This expanding diversity in text generation mechanisms necessitates robust, standardized benchmarking protocols to evaluate decoding performance systematically.

The conventional benchmarking framework for open-ended text generation typically employs a curated benchmark \textit{suite}, such as WikiText \citep{merity2016pointer} or WikiNews\footnote{\href{https://github.com/XiangLi1999/ContrastiveDecoding}{\url{www.wikinews.org} and utilizing benchmark split by \cite{li2023contrastive}}}, comprising a representative set of prompts to comparatively assess multiple decoding strategies through automated quality metrics and/or human evaluations. This benchmarking serves two critical functions: For practitioners, it provides empirical guidance for selecting optimal decoding strategies tailored to specific applications. For researchers developing novel decoding algorithms, it offers quantitative insights into the relative performance of existing methods and enables precise measurement of improvements introduced by new approaches, as further discussed in Section~\ref{sec:related_work}. Despite their importance, current benchmarking methodologies for evaluating decoding strategies in open-ended text generation remain constrained by three fundamental unresolved challenges that compromise both their utility and reliability.

\begin{figure}
    \centering
    \includegraphics[width=1.0\textwidth]{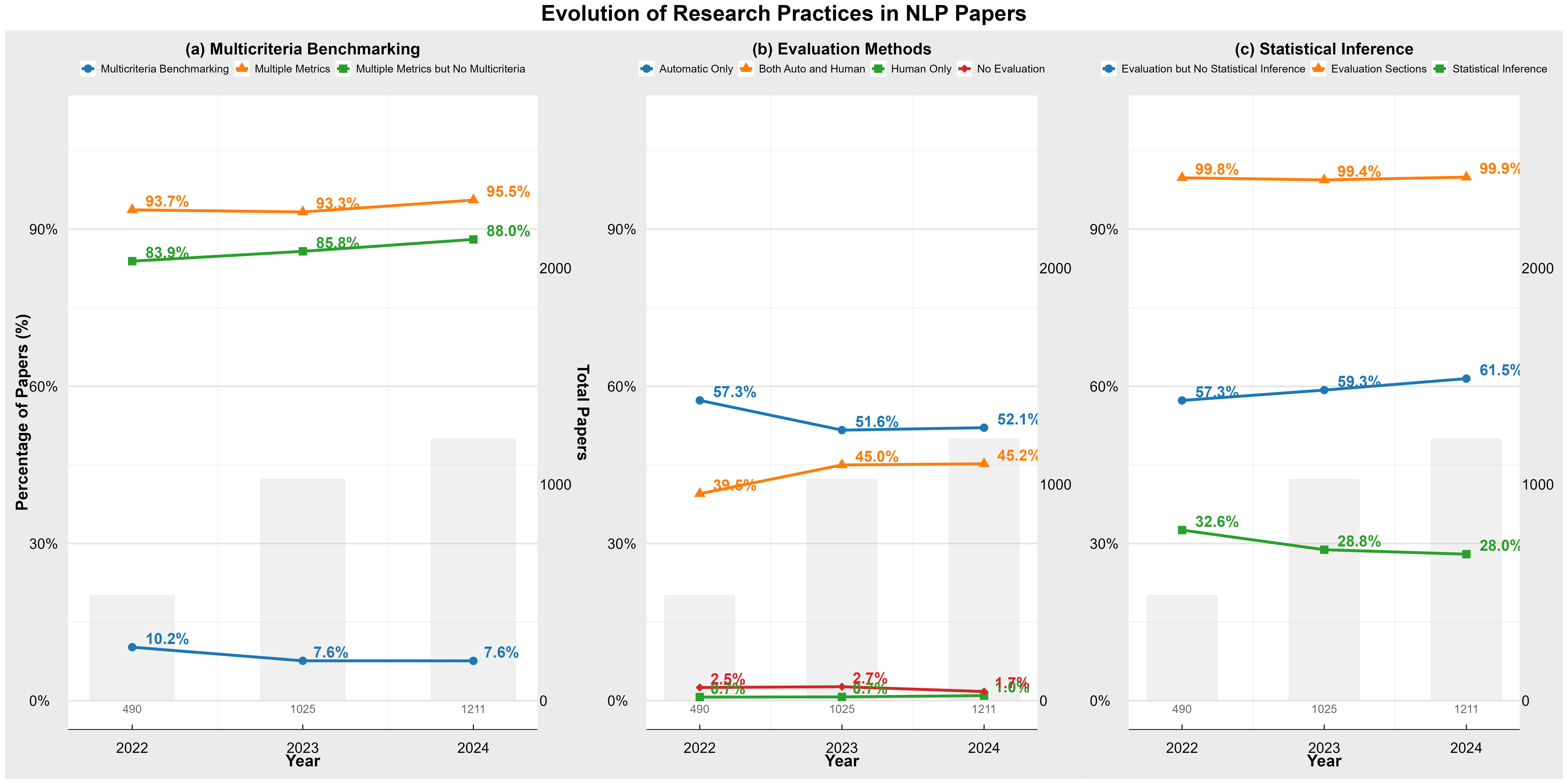}
\caption{Analysis of text generation research trends (arXiv, 2022-2024): (a) Evolution of multicriteria benchmarking showing individual metric evaluation (orange) versus true multicriteria approaches (blue). (b) Distribution of evaluation methodologies across automatic (blue), human (green), combined (orange), and no evaluation (red) approaches. (c) Adoption of statistical inference methods (green) compared to evaluations without statistical validation (blue).}
    \label{fig:research_evolution}
\end{figure}

\paragraph{(I) Reliance on Single Metrics.} First, conventional benchmarking methods typically rely on a single metric to assess the quality of LLM-generated texts (cf. Figure \ref{fig:research_evolution}(a)). However, \textit{text quality} is inherently multidimensional, making its reduction to a single value a fundamental oversimplification. This multidimensionality is evidenced by the diverse range of established automatic text evaluation metrics\footnote{We use the term "automatic" to describe metrics that can be calculated directly from the generated text without human evaluation. Formulas and definitions for these metrics are provided in Appendix \ref{a: autom_metrics}.}, including perplexity (\texttt{ppl}), diversity (\texttt{div}), and coherence (\texttt{coh}), commonly employed in decoding research \citep[see, e.g.,][]{holtzman2019curious,su2022contrastive,arias2025betteropenendedtextgeneration}. Each metric is assumed to capture a distinct aspect of the latent construct \textit{text quality}: \texttt{div} quantifies lexical repetition, while \texttt{coh} measures similarity between the prompt and the generated text. Even human judgments of text quality exhibit substantial interpersonal variation, further confirming the inherently complex, multidimensional nature of quality assessment. Despite this evident limitation of single metrics and naive aggregations thereof (Q*text, see Appendix \ref{a:qtext_technical}), Figure~\ref{fig:research_evolution}(a) reveals an increase in papers evaluating multiple metrics individually (from 83.9\% in 2022 to 88.0\% in 2024), while true multicriteria benchmarking approaches remain rare and are even declining (from 10.2\% in 2022 to 7.6\% in 2024). This suggests that although researchers acknowledge the importance of multiple evaluation dimensions, comprehensive multicriteria approaches remain significantly underutilized in practice.

\paragraph{(II) Integrating Human Evaluation.} Second, the conventional restriction to a single quality metric overlooks the substantial benefits of combining automatic and human evaluations. Human assessments of LLM-generated text are widely accepted as the gold standard for overall quality measurement, while automatic metrics provide precise quantification of specific attributes \textendash, such as diversity, which measures lexical richness rather than merely favoring~\say{safe} responses \citep{tevet-berant-2021-evaluating, li2015diversity}. Figure \ref{fig:research_evolution}(b) documents the evolution of evaluation methodologies, revealing that while automatic evaluation remains dominant (52-57\% of papers), there is a clear trend toward combined automatic-human evaluation methods (increasing from 39.5\% in 2022 to 45.2\% in 2024). This shift indicates growing recognition that complementary assessment approaches yield more comprehensive quality measurements. Nevertheless, substantial challenges persist: beyond the methodological difficulties in standardizing human assessment protocols, the fundamental differences in measurement scales pose a critical problem. Automatic metrics typically provide cardinal measurements, enabling quantitative comparisons (e.g., \say{Decoding strategy A is \textit{twice as effective as} strategy B}), whereas human evaluations generally yield ordinal data (often through Likert scales), indicating relative quality (e.g., \say{Strategy A is \textit{better than} strategy B}) without quantifying the magnitude of difference. These distinct measurement properties \textendash, cardinal versus ordinal \textendash, require careful methodological integration in benchmarking approaches.

\paragraph{(III + IV) Generalizing Beyond the Sample.} Third, most recent evaluation approaches remain purely descriptive, lacking statistical rigor. Figure \ref{fig:research_evolution}(c) reveals a concerning trend: the proportion of papers employing statistical inference methods has declined from 32.6\% to 28.0\%, while evaluations without statistical validation have increased from 57.3\% to 61.5\%. This pattern indicates that despite the increasing prevalence of evaluation overall, many publications fail to implement proper statistical validation through significance testing or confidence intervals. To address this, robust benchmarking approaches need to incorporate inferential statistics, acknowledging that text generations represent only a sample from a broader population. These statistical methods necessarily rely on specific assumptions about the sampling process generating both prompts and evaluations. Most commonly, they assume an independent and identically distributed (i.i.d.) framework, wherein prompts and evaluations derive from the same distribution and maintain mutual independence. In practice, however, human evaluations often violate independence assumptions; for instance, exposure to an exceptionally high-quality text may systematically influence subsequent judgments. By explicitly accounting for these statistical dependencies, benchmarking methods can provide more generalizable insights that transcend the limitations inherent to any single benchmark suite.

\paragraph{Contributions.} In this study, we address the fundamental challenges in benchmarking LLM-generated text. When developing a novel decoding strategy, researchers inevitably face a critical question: Does this new method produce text of comparable or superior quality to established approaches? If we can demonstrate that a new method performs at least as well as existing methods, it warrants further investigation and represents a meaningful scientific contribution. To systematically evaluate such questions, we introduce a rigorous benchmarking framework that addresses three critical limitations in current evaluation paradigms. Our GSD approach ...
\begin{enumerate}
    \item[\textbf{(I)}] ... incorporates multiple quality metrics simultaneously, capturing the inherent multidimensionality of text quality.
    \item[\textbf{(II)}]... integrates both human and automatic evaluations while preserving their distinct measurement scales.
    \item[\textbf{(III)}] ... provides inferential statements, thus extending beyond limited descriptive evaluations.
    \item[\textbf{(IV)}] ... is able to quantify the robustness of these inferential statements under potential deviations from the i.i.d.~assumption. 
\end{enumerate}

We present the GSD-front methodology and its corresponding statistical test to determine whether a new decoding strategy produces text of sufficiently high quality to merit inclusion in future research. Specifically, we assess whether its quality is competitive with existing strategies while addressing the key challenges outlined above. Our approach is grounded on recent developments in decision theory \citep{jsa2018,jbas2022}, which provides a framework for comparing uncertain outcomes using statistical evidence. By combining ideas from robust statistics \citep{Huber:1981} and imprecise probabilities \citep{w1991}, we enhance the robustness of our statistical methods, particularly when the assumption of i.i.d. prompts is not fully satisfied.

We validate our approach using a curated selection of prompts from the WikiText and WikiNews datasets (see Section~\ref{sec:evaluation_results}), which also serve as illustrative examples to elucidate our theoretical framework. Rather than comparing a novel method against established ones, we investigate whether human-written completions remain potentially superior to automated text generation approaches. To this end, we evaluate five distinct decoding strategies (detailed in Table \ref{tab:decoding_strategies}): beam search \citep{Freitag_2017}, contrastive search \citep{su2022contrastive}, sampling with temperature \citep{ackley1985learning}, top-\( k \) sampling \citep{fan2018hierarchical}, and nucleus (top-\( p \)) sampling \citep{holtzman2019curious}, benchmarking each against original human completions. Our evaluation incorporates Q*Text, an automatic cardinal metric that aggregates perplexity, diversity, and coherence \citep{arias2025betteropenendedtextgeneration}, alongside human evaluations using an ordinal scale (see Table \ref{tab:quality-scale} and Figure \ref{fig:evaluation} in Appendix \ref{a:human_evaluation}). Our statistical analysis reveals that, as expected, human-written completions still enhance—or at minimum do not diminish—text quality compared to current algorithmic decoding strategies. We make our codebase, algorithm documentation, evaluation results and data publicly available\footnote{All resources, including computation procedures and complexity, are available at: 
\\\url{https://github.com/hannahblo/Statistical_Multicriteria_Evaluation_of_LLM-Generated_Text}}.

\section{Generalized Stochastic Dominance and the GSD-Front}
\label{gsd_dominance_front}

To address the challenges outlined in Section \ref{sec:intro}, we propose a framework based on generalized stochastic dominance (GSD). Rather than imposing a complete ranking—which would require potentially unjustified assumptions about the relative importance of quality dimensions—our framework identifies a minimal set of non-dominated strategies. This set, which we term the "GSD-front," represents potentially optimal choices under varying preference structures. In this section, we present the theoretical foundation of our approach.

\subsection{Generalized Stochastic Dominance}
GSD has received quite some interest recently \citep[e.g.][]{mdai,jansen2023statistical,jansen2024statistical,jansen2025contributions}. We here propose to adapt the GSD-front, introduced in \citet{jansen2024statistical} for the purpose of classifier selection, as a method for comparing decoding strategies with respect to several quality metrics simultaneously. The basic idea is quite natural: We first utilize the multidimensional order structure spanned by the quality metrics for defining a \textit{partial expectation ranking} among the decoding strategies under consideration. In our application these are Q*text and the two human evaluations. Afterwards, we select the non-strictly dominated strategies under this order to be included in the GSD-front. In our application, we consider all decoding strategies together with the human completion and select those which are not strictly dominated (i.e. systematically worse) than any of the others. Hence, if the human completion lies in the GSD-front it is not dominated by any of the other five automatic decoding strategies and therefore can potentially produce higher quality text than those in certain situations. Note that the decoding strategies in the GSD-front are incomparable to each other (GSD is a partial order) and, in general, no unique best decoding strategy will be obtained in this way. However, it can be argued that the GSD-front represents the smallest set of incomparable strategies that can be obtained without (potentially hard-to-justify) additional assumptions about the weighting of the quality metrics since it incorporates the entire information encoded in both the (empirical) distribution of the prompts and the order structure induced by the metrics. 
\subsection{Technical Setup} \label{tech_set}
Assume we are given a finite set $\mathcal{S}$ of decoding strategies, a universe $\mathcal{P}$ of prompts, and $n$ quality metrics $\phi_1 , \dots , \phi_n:\mathcal{S}\times \mathcal{P} \to [0,1]$. For every  $i \in \{1, \dots , n\}$, $S \in \mathcal{S}$, and $P \in \mathcal{P}$, the value $\phi_i(S,P)$ describes quality of the completion obtained by applying strategy $S$ to prompt $P$ with respect to the metric $\phi_i$ (where higher values indicate better quality). 
To clearly distinguish between ordinal and cardinal evaluations, we assume that, for $0 \leq z \leq n$, the metrics $\phi_1, \dots , \phi_z$ are of cardinal scale (differences may be interpreted), while the remaining ones are purely ordinal (differences are meaningless apart from the sign). Given this setup, we then consider the combined multidimensional metric 
\begin{equation}
\Phi:=(\phi_1 , \dots , \phi_n):\mathcal{S}\times \mathcal{P} \to [0,1]^n.
\end{equation}
We define two binary relations associated with the range $\Phi(\mathcal{S}\times \mathcal{P})$ of $\Phi$, i.e., the set of quality vectors spanned by the considered quality metrics. The first of these relations captures the \textit{ordinal information} encoded in the multidimensional quality evaluations, while the second relation captures the \textit{cardinal} part of the information:  

\noindent \textbf{Ordinal Information:} For any pair of quality vectors $x=\Phi(S,P), y=\Phi(S',P')$, we set
\begin{equation}
(x,y) \in R_1 ~~:\Leftrightarrow ~~ \forall i: \phi_i(S,P) \geq \phi_i(S',P').
\end{equation}
In other words, if the two quality vectors $x$ and $y$ (defined as above) are in relation with respect to $R_1$, this means that the completion of $P$ by $S$ is judged at least as good as the completion of  $P'$ by $S'$ by any of the considered quality metrics.

\noindent \textbf{Cardinal Information:} For any quadruple of quality vectors $t=\Phi(S,P), u=\Phi(S',P'),v=\Phi(L,G), w=\Phi(L',G')$, such that $(t,u),(v,w) \in R_1$, we set
\begin{equation*}
    ((t,u),(v,w)) \in R_2 ~~:\Leftrightarrow ~~\forall i\leq z ~ \forall j> z
\end{equation*}
\begin{equation*}
    \phi_i(S,P) - \phi_i(S',P')\geq\phi_i(L,G) - \phi_i(L',G')~~\wedge ~~\phi_j(S,P) \geq\phi_j(L,G) \geq \phi_j(L',G') \geq \phi_j(S',P')
\end{equation*}
In other words, if two ordered pairs of quality vectors $(t,u)$ and $(v,w)$ (as defined above) are in relation with respect to $R_2$, this means that whenever the ordinal components of the latter quality vectors are bounded (from above and below) by the ordinal components of the further quality vectors, we can compare intensity of preference between quality vectors by comparing their differences in the cardinal components.

\noindent These considerations leave us with a partially-cardinal scaled order structure 
\begin{equation}
\mathbb{P}=(\Phi(\mathcal{S}\times \mathcal{P}),R_1 , R_2)    
\end{equation}
on the basis of which we intend to analyze the performance of the decoding strategies under consideration. Note that this structure encodes exactly that quality information that can be obtained from the data without additional assumptions about the weighting of the involved quality metrics. To ease working with $\mathbb{P}$, we replace it by the set of \textit{utility representations} respecting its structure. Intuitively, each of those utility functions can then be interpreted as a candidate measurement scale (or, in other words, a potential cardinal completion) that is compatible with the information encoded in $\mathbb{P}$, i.e., the information arising from the mixed-scaled multidimensional quality evaluations across the considered combination of decoding strategies and prompts in the set  $\mathcal{S} \times
\mathcal{P}$.  

\textbf{Utility Representation:} Call a function $u:\Phi(\mathcal{S}\times \mathcal{P}) \to \mathbb{R}$ compatible with (or utility representation of) $\mathbb{P}$, whenever for all $(x,y)\in R_1$ and all $((r,t),(v,w)) \in R_2$ it holds that: 
\begin{equation}
    u(x)\geq u(y) ~~~ \wedge ~~~u(r)-u(t)\geq u(v)-u(w)
\end{equation}
We then denote by $\mathcal{U}_{\mathbb{P}}$ the set of all (bounded and measurable) functions that are compatible with $\mathbb{P}$. This set then captures all the relevant information encoded in the structure $\mathbb{P}$, however, is much more accessible for a meaningful analysis.

\noindent The set $\mathcal{U}_{\mathbb{P}}$ of utility representations obtained $\mathbb{P}$ now forms the basis for the generalized stochastic dominance (GSD) relation among the decoding strategies under consideration.  Moreover, note that defining the GSD-relation on the set $\mathcal{S}$ requires to assume that the prompts in $\mathcal{P}$ are generated randomly according to some probability measure $\pi$ (note that for our actual analysis this will be replaced by its empirical analog). 

\textbf{Generalized Stochastic Dominance:} We say that decoding strategy $S$ GSD-dominates decoding stategy $S'$, denoted by $S \succsim S'$, whenever it holds that:
\begin{equation}
\forall u \in \mathcal{U}_{\mathbb{P}}:~~\mathbb{E}_{\pi}(u \circ \Phi(S, \cdot)) \geq \mathbb{E}_{\pi}(u \circ \Phi(S', \cdot))    
\end{equation}
Note that the GSD-relation $\succsim$ is not complete, i.e., in general, there will exist decoding strategies that are incomparable with respect to GSD.

\noindent The last step is adapting the GSD-front to the comparison of decoding strategies.  Again, this can be done straightforwardly: We simply collect the non-strictly dominated strategies with respect to the GSD-relation $\succsim$ that we adapted to this context in the previous step. 

\textbf{GSD-Front:} Formally, the GSD-front is thus given by
\begin{equation}
\text{gsd}(\mathcal{S})=\{S \in \mathcal{S}:~ \nexists S' \in \mathcal{S} \text{ s.t. } S' \succ S\},    
\end{equation}
where $\succ$ denotes the strict part of $\succsim$.

\noindent  Reflecting that $\text{gsd}(\mathcal{S})$ will, in general, be inaccessible since the true law $\pi$ is unknown, in practice we will often have to make do with its empirical version, i.e.,~the set $\text{gsd}_{emp}(\mathcal{S})$ that is obtained by replacing all population-based expressions in $\text{gsd}(\mathcal{S})$ by their empirical analogs. Note, however, that $\text{gsd}_{emp}(\mathcal{S})$ makes a mere descriptive statement on the relation of the decoding strategies. 

\textbf{Empirical GSD-Front and (Robust) Statistical Testing:} To move to inferential guarantees, a statistical test for the pair
\begin{equation}
    H_0: S \notin \text{gsd}(\mathcal{S})~~\text{\textbf{vs.}}~~ H_1: S \in \text{gsd}(\mathcal{S}) 
    \label{hypotheses}
\end{equation}
is desirable: If $H_0$ can be rejected at a level $\alpha$ using an appropriate test, there is significant evidence that the decoding strategy $S$ is competitive with the strategies in $\mathcal{S}\setminus \{S\}$ in certain situations \textit{across the population of prompts} and, accordingly, should be further considered. But how can an appropriate statistical test be constructed? \citet{jansen2024statistical} demonstrate that (under mild assumptions that are met in our situation) a valid and consistent test is indeed reachable (by using an adapted permutation testing scheme). Furthermore, they show that this statistical test can be robustified to samples (slightly) deviating from the usual i.i.d. assumption by relying on techniques originating in robust statistics. In particular, their techniques allow us to analyze the $p$-value of a test decision for the pair $(H_0,H_1)$ as a function of the \textit{contamination size} of the underlying sample of prompts, i.e., the share of prompts stemming from some arbitrary distribution. In our context, such robustification seems particularly relevant: Especially when a large number of completions are evaluated by humans in a short period, certain implicit dependency structures are often difficult to avoid.

\noindent For interpreting the test results in Figure \ref{fig:results}, it is important to note that the test proposed in \citet{jansen2024statistical} for the hypothesis pair $(H_0,H_1)$ consists of a series of \textit{pairwise comparison tests} of strategies regarding their GSD relation. To be precise, the strategy $S$ is tested against all strategies in $\mathcal{S}\setminus \{S\}$ and $H_0$ is rejected if all these sub-tests reject their respective null hypotheses. The test statistic used for each of those pairwise comparison ($S$ versus $S'$) tests is based on the empirical version of 
\begin{equation}
    D(S,S'):= \inf_{u \in \mathcal{U}_{\mathbb{P}}} \Bigl\{\mathbb{E}_{\pi}(u \circ \Phi(S, \cdot)) - \mathbb{E}_{\pi}(u \circ \Phi(S', \cdot))\Bigr\},
    \label{test_stat}
\end{equation}
i.e., the expression arising from $D(S,S')$ by exchanging all population concepts by empirical analogs.

\section{Application: Automatic and Human Quality Evaluation}\label{sec:evaluation_results}

In this section, we present a application where we investigate whether human text generation potentially still offers superior quality compared to alternative automatic decoding strategies. Please note that our method can be applied to any set of competing decoding methods. This investigation addresses the three key benchmarking challenges outlined in Section \ref{sec:intro}: analyzing multiple quality metrics with different measurement scales simultaneously (Challenges I and II), and quantifying the robustness of inferential statements under potential deviations from the i.i.d.~sampling assumption (Challenge III and IV). The latter arises specifically from potential dependencies in human evaluations and the use of prompts from two distinct datasets.

\subsection{Experimental setup}
\label{sec:experimental_setup}

To demonstrate our method's application, we employed the Qwen 2.5 - 7B model \citep{yang2024qwen2technicalreport} with prompts from Wikitext \citep{merity2016pointer} and Wikinews\footnote{\href{www.wikinews.org}{Wikinews from \url{http:\\www.wikinews.org}}}, incorporating diverse and factually-grounded contexts. Our sample encompassed 300 text generations—50 prompts with six continuations each: one human-written (H) and five generated using different decoding strategies. These strategies included both deterministic methods: beam search (BS) and contrastive search (CS), as well as stochastic approaches: temperature sampling (TS), top-$k$ sampling ($T_{k}$), and nucleus top-$p$ sampling ($T_{p}$). Detailed descriptions of these strategies appear in Table~\ref{tab:decoding_strategies} in Appendix~\ref{a:decoding_strategies}.

Our evaluation framework integrated both human assessments and automated metrics. Human evaluators rated text quality on a 5-point Likert scale ranging from 1 (low quality) to 5 (high quality), see Table \ref{tab:quality-scale}, following instructions detailed in Appendix \ref{a:human_evaluation}. Though the evaluators were authors of this paper, we implemented a blind evaluation protocol where they scored texts without knowledge of their source or the decoding strategy used, minimizing potential biases \citep{belz-etal-2020-disentangling}. We complemented these subjective judgments with cardinal Q*Text scores that synthesize generation perplexity, diversity, and coherence metrics as established by \citet{arias2025betteropenendedtextgeneration}. For a technical overview about this metric and its components, we refer to Section~\ref{a: autom_metrics}.

\begin{figure}[ht]
\begin{subfigure}{.5\textwidth}
  \centering
  \includegraphics[width=0.95\linewidth]{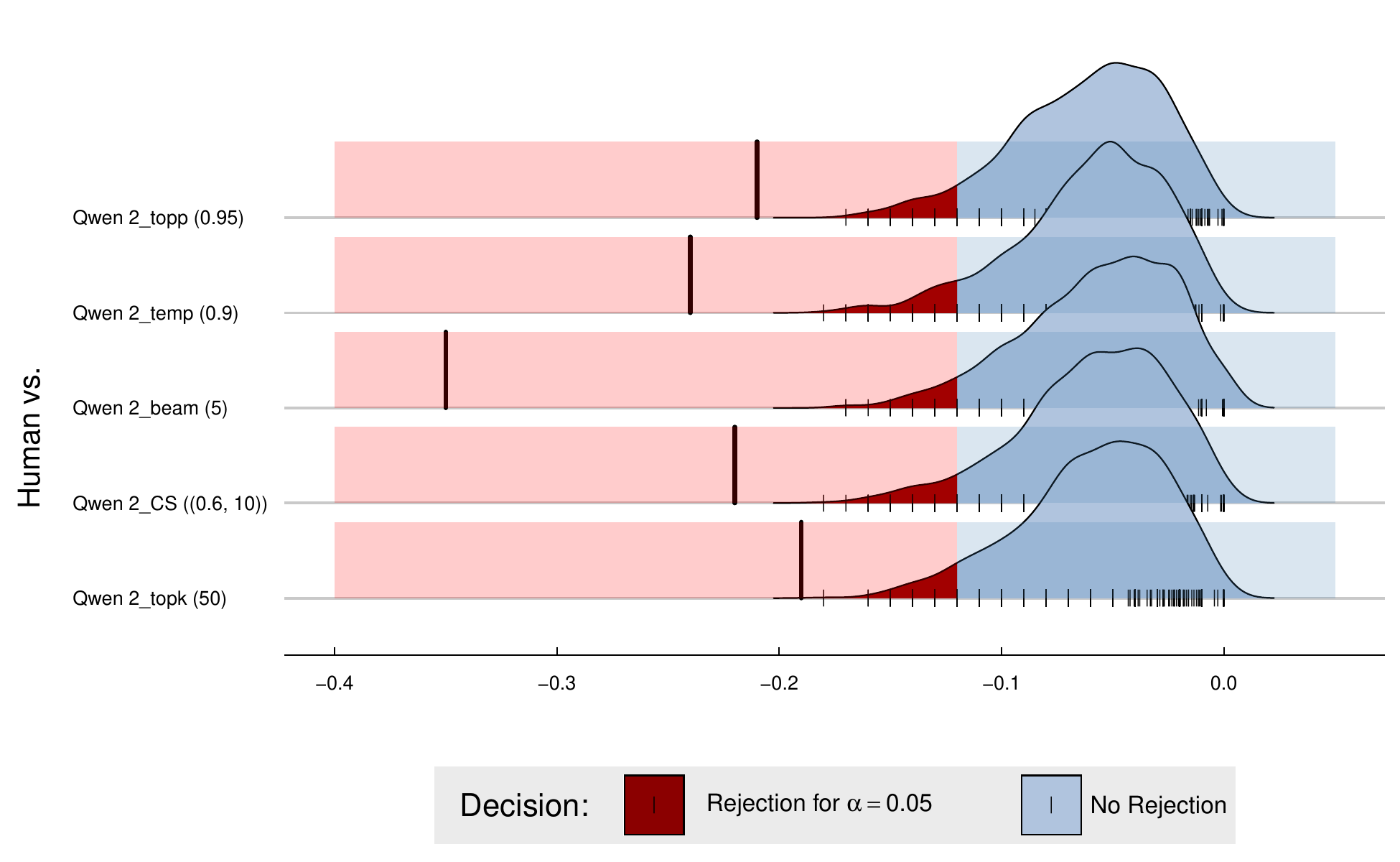}
\end{subfigure}%
\begin{subfigure}{.5\textwidth}
  \centering
  \includegraphics[width=0.95\linewidth]{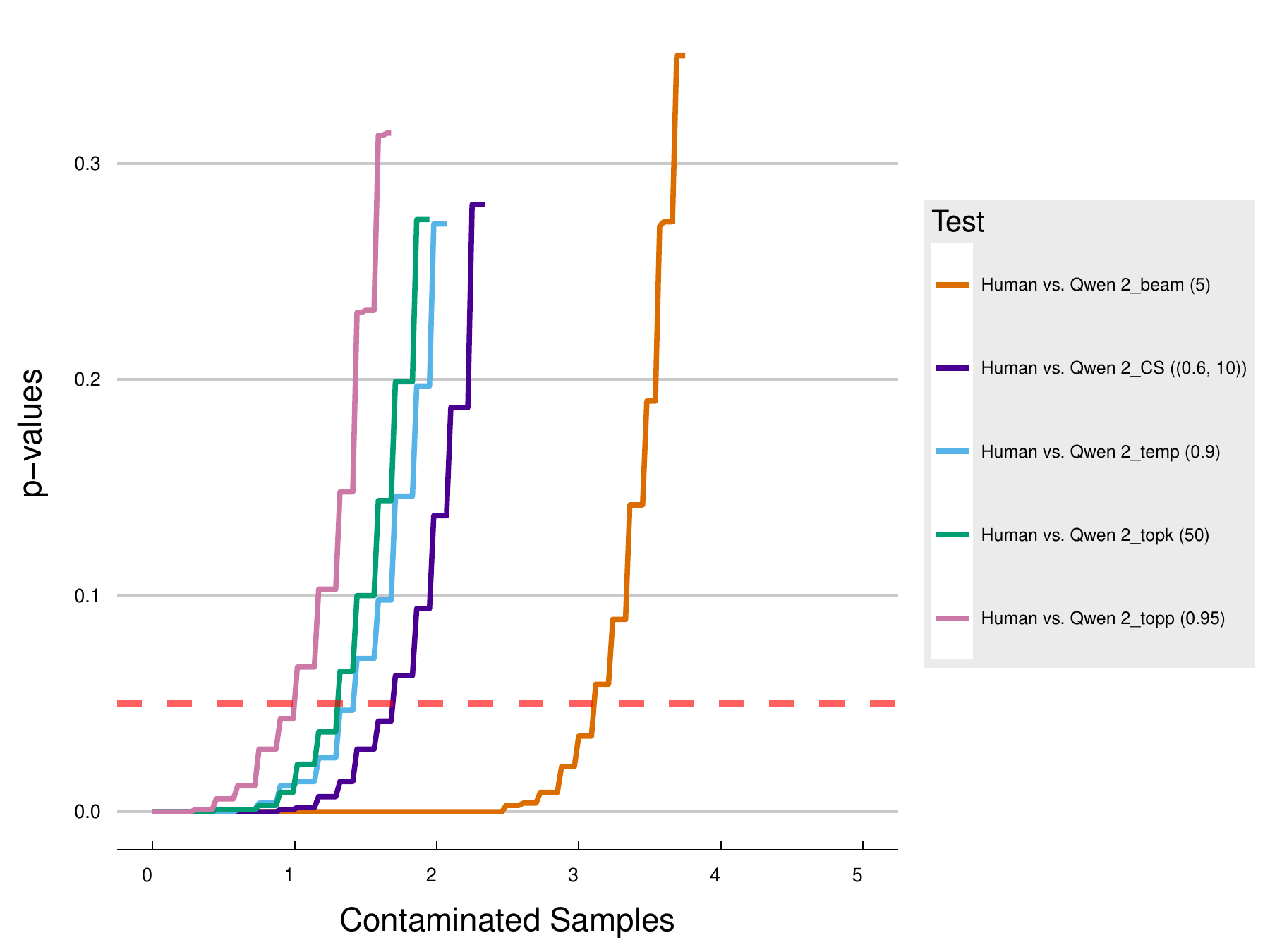}
\end{subfigure}
\caption{Left: Empirical densities of resampled test statistics for pairwise GSD-comparisons between \textit{human} and five decoding strategies, using prompts from Wikinews and Wikitext. Vertical markers indicate the observed test statistic values, with the rejection threshold ($\alpha = 0.05$) marked in red.
Right: Assessment of i.i.d.~assumption violations for the same pairwise GSD-comparisons between \textit{human} and five decoding strategies. The plot shows computed p-values with the significance threshold ($\alpha = 0.05$) indicated by the red horizontal line.}
\label{fig:results}
\end{figure}

\subsection{Representation within the GSD framework}
\label{sec:application}

As previously stated, we compare human-generated text completion (H) to five decoding strategies: BS, CS, TS, T$_k$, and T$_p$ based on prompts from the WikiText/WikiNews benchmark suites, (see Section~\ref{sec:experimental_setup}). The set of decoding strategies is defined as:
\begin{equation}
    \mathcal{S} = \{\text{H, BS, CS, TS, T$_k$,  T$_p$}\},
\end{equation}
whereas the set $ \mathcal{P}$ represents the underlying population from which the prompts in WikiText/WikiNews are sampled. We evaluate text quality using three metrics: Q*Text (denoted as $\phi_1$), which provides cardinal quality assessments, while metrics $\phi_2$ and $\phi_3$ are ordinal and based on evaluations from two of the paper's authors (hence, the number of cardinal dimensions is $z = 1$).

Following Section~\ref{gsd_dominance_front}, we define two ranking relations on the set of quality vectors spanned by our (mixed-scaled) three-dimensional performance metric $\Phi=(\phi_1,\phi_2,\phi_3)$: $R_1$ is a (partial) order that ranks the quality vectors associated with our metric $\Phi$ based on the ordinal evaluations ($\phi_1, \phi_3$) of the completions for the prompts from $ \mathcal{P} $ as well as the cardinal evaluation from the automated metric Q*Text ($\phi_1$). 
$R_2$ is defined as a relation capturing the difference in intensity between pairs of quality vectors associated with the multdimensional metric $\Phi$. As described in detail in Section \ref{gsd_dominance_front}, we can use these two relations $R_1$ and $R_2$ to obtain a ranking of the decoding strategies in $\mathcal{S}$ by applying (empirical) generalized stochastic dominance.

We use empirical GSD to represent these rankings and assess each decoding strategy's performance. The empirical GSD-front consists of all decoding strategies that are not strictly outperformed, i.e. \textit{dominated} across all three metrics by others. This allows us to investigate whether human text completion enhances the quality of generated text compared to the five decoding strategies. For a comprehensive description of the quality assessment criteria, we refer to Table \ref{tab:quality-scale} in Appendix \ref{a:human_evaluation}.

\subsection{Results}
\label{sec:results}

To examine whether human text generations can potentially improve on completion quality compared to the aforementioned automatic strategies (see Section \ref{sec:experimental_setup}), we conduct the statistical test for the GSD-front as described in the paragraph following Equation \eqref{hypotheses} in Section~\ref{tech_set} based on the following specification of the null hypothesis:
\begin{align*}
    H_0: \text{H} \not\in \text{gsd}(\{ \text{H, BS, CS, TS, T$_k$, T$_p$} \})
\end{align*}
at a significance level of $\alpha = 0.05$. As described in Section \ref{gsd_dominance_front}, to test the hypotheses pair $(H_0,H_1)$, we perform statistical tests for five auxiliary null hypotheses, each corresponding to a pairwise GSD-dominance comparison between human text completion (H) and one of the automatic text completion strategies BS, CS, TS, T$_k$, and T$_p$ (the detailed testing schemes for those auxiliary tests can be found in \citet[A.2.2]{jansen2024statistical}). The distribution of the resampled pairwise test statistics, i.e., the empirical versions of $D(H,S),$ where $S\in \mathcal{S}\setminus \{H\}$ (see Equation \eqref{test_stat}), is illustrated in Figure~\ref{fig:results} (left). It demonstrates that the pairwise tests are significant across all five comparisons. Consequently, we conclude that human completion is not significantly outperformed by any of the automatic decoding strategies in the set $\mathcal{S}$ and, therefore, can be assumed to lie in the GSD-front gsd$(\mathcal{S})$ of the considered set of decoding strategies $\mathcal{S}$ at level $\alpha = 0.05$.

In other words, \textit{we find no evidence to suggest that human completion is redundant—the considered decoding strategies have not yet reached a quality level where human completion offers no additional value}. By incorporating statistical inference, we have moved beyond mere descriptive analysis to an inductive analysis that extends beyond the benchmark suites WikiText/WikiNews.

As emphasized in Section~\ref{sec:intro}, robust inference analysis should also account for potential deviations from the i.i.d. assumption (see the last paragraph of Section \ref{gsd_dominance_front} for further details). While recommendable in general, this is especially true for applications like the one at hand, where the assumption of identical distributions is questionable because of the sampling scheme of the prompts (two different suites) and the independence assumption is questionable because of the way the text evaluations of the human evaluators are obtained (potential learning effects during the evaluation process). Therefore, in Figure~\ref{fig:results} (right), we examine the robustness of our test decision under contamination of the benchmark suite, specifically considering deviations from the i.i.d.~assumption. The figure displays the $p$-values of all five (significant) auxiliary tests as functions of the contamination size of the benchmark suite, i.e., the degree of deviation from the i.i.d. sampling assumption of the prompts.  We see that the pairwise comparison results remain significant as long as at most 1 (for the pink, the green, the blue, and the purple line) or 3 (for the yellow line) prompt(s) deviate(s) from this assumption. Since all five pairwise comparisons must be significant to reject the null hypothesis above, we conclude that, at most, one prompt can deviate (and stem from some arbitrary distribution) while maintaining the statistical significance of our test decision. 

Beyond the specific results for the concrete application, this casts an interesting light on reliable statistical statements in benchmark studies in general: Since such statements (especially those of inferential nature) depend heavily on idealizing assumptions about the analyzed benchmark suites, it is all the more important that benchmark suites are curated according to appropriate standards.

\section{Related Work}
\label{sec:related_work}

Benchmarks constitute a fundamental pillar in applied machine learning, serving not merely as selection guides for practitioners but as critical platforms for methodological innovation and validation \citep{ye2024benchmarking,hu2020open,thiyagalingam2022scientific,kirk2024prism, zhang2024inherent,shirali2023theory,ott2022mapping,zhang2020machine,bongratz2024choose,roelofs2019meta,vanschoren2014openml}. Their instrumental role in accelerating scientific progress is evidenced by numerous studies that systematically leverage benchmarks to demonstrate the efficacy of novel approaches \citep{shi2022minimax,MEYER2003169,bordini2024self,Hothorn,ehl2012,mptbw2015,rodemann2024explaining,rodemann2023all,rodemann2023approximately,rodemann2022levelwise,dietrich2024semi,nalenz2024learning,caprio2023credal,rafailov2023direct,ziomek2024time,av2025accelerated,xu2024principled,adachi2025bayesian}.

Recent investigations have revealed critical challenges regarding the replicability and reliability of benchmark outcomes. \cite{berrar2024estimating} demonstrate that performance improvements observed in isolated experiments frequently fail to replicate across multiple trials, while \cite{madaan2024quantifying} establish that seemingly minor factors—including random initialization and data sampling variations—can fundamentally alter performance rankings \citep{white2024livebench,zhou2023don}. These findings, coupled with comprehensive surveys within the NLP community \citep{dror2018guide}, underscore the imperative for more robust, transparent, and statistically rigorous evaluation methodologies.

To address these methodological shortcomings, researchers have increasingly developed statistically rigorous frameworks for detecting significant performance differences between competing approaches, explicitly acknowledging that benchmark datasets represent finite samples drawn from much larger underlying populations \citep{demvsar2006statistical,Garca2008AnEO,GARCIA20102044,benavoli2017time}. This movement has gained further momentum through the emergence of multi-criteria and multi-task benchmarking paradigms \citep{cruz2024evaluating,zhanginherent,kohli2024towards,jansen2024statistical,jansen2023robust,jansen2023statistical,rodemann2024partial,blocher2024comparing}. Across diverse domains—from predictive machine learning, where accuracy must be balanced against computational efficiency \citep{koch2015efficient,jansen2024statistical}, to optimization tasks requiring simultaneous optimization of performance and speed \citep{schneider2018deepobs}—the need to systematically reconcile competing performance indicators has become increasingly evident. 

This paradigm is equally applicable to neural text generation, where researchers have proposed numerous metrics to assess different aspects of output quality \citep{hashimoto2019unifyinghumanstatisticalevaluation,alihosseini-etal-2019-jointly,pillutla2021mauve,celikyilmaz2021evaluationtextgenerationsurvey,su2022empirical,su2022contrastive,gao2022simcse,becker2024textgenerationsystematicliterature,garces-arias-etal-2025-decoding}. Each metric evaluates a distinct dimension: diversity quantifies lexical richness through n-gram repetition rates, MAUVE measures the distributional similarity between generated and reference texts, coherence calculates the log-likelihood of generated text given its prompt, and generation perplexity assesses text predictability according to the model's distribution \citep{10.1121/1.2016299}. Relying on any single metric, however, provides an incomplete evaluation. Optimizing solely for coherence often results in repetitive outputs, a phenomenon referred to as \textit{degeneration}, while maximizing diversity may compromise semantic integrity \citep{lee2022factuality,holtzman2019curious}. Despite this, the application of multicriteria benchmarking and statistical inference has declined since 2022, compare Figures \ref{fig:research_evolution}(a) and (c).

Complicating matters further, a notable discrepancy often emerges between automatic metrics and human evaluations \citep{su2022empirical,arias2024adaptivecontrastivesearchuncertaintyguided,rodemann2025statisticalcaseempiricalhumanai}; in response, integrated evaluation frameworks have been developed. A notable example, HUSE \citep{hashimoto2019unifyinghumanstatisticalevaluation},  combines human judgments with model-based probabilities to jointly assess quality and diversity. In many cases, automatic metrics are cardinally scaled, while human evaluations are typically available as ordinal data, see challenge (II) in Section~\ref{sec:intro}. In this work, we propose a multicriteria benchmarking framework that utilizes both cardinal \textit{and} ordinal information in an information-efficient way, see Sections~\ref{sec:intro} and~\ref{gsd_dominance_front}.

Recent benchmarking advances for neural text generation include the multicriteria framework by \cite{arias2025betteropenendedtextgeneration}, which combines diverse ranking strategies. Their approach uses the extended Bradley-Terry model to convert direct comparisons into pairwise worth parameters, creating a strict total ordering of decoding methods. To address mutual incomparability, they applied union-free generic (ufg) depth, enabling partial orders that capture preference structures without forcing complete rankings. They also introduced \textit{Q*Text}—a single metric calculated as the harmonic mean of penalized coherence, diversity, and perplexity—to integrate multiple evaluation dimensions.


\section{Conclusion}
\label{sec:conclusion}

We introduced a framework based on Generalized Stochastic Dominance (GSD) that addresses three critical limitations in current methodologies for evaluating LLM-generated text: (1) the inadequacy of single-metric assessment, (2) the incompatibility between cardinal automatic metrics and ordinal human judgments, and (3) the absence of robust statistical guarantees. The GSD-front approach integrates multiple quality dimensions while preserving their distinct measurement scales and enables quantifying the robustness of inference under potential deviations from i.i.d. assumptions. To validate this framework, we conducted a comparative analysis of five common decoding strategies against human-written text, though the method generalizes to any set of generation approaches. The GSD-front enables statistically sound multicriteria evaluation without requiring arbitrary metric weighting or compromising measurement scale integrity. By incorporating techniques from robust statistics, our approach extends beyond descriptive benchmark analysis to provide inferential guarantees that account for potential dependencies in human evaluations. This advancement provides researchers and practitioners with a more rigorous methodology for evaluating text generation systems. Future work could extend the GSD approach to other generation tasks such as summarization and translation, investigate additional quality dimensions, and further enhance statistical robustness for complex evaluation dependencies.

\paragraph{Limitations.}
Despite the strengths of our proposed framework, several limitations should be acknowledged. First, our experimental validation focused primarily on benchmarking human text continuations with LLM-generated text in an open-ended text generation task. While this provided a suitable context for demonstrating our framework, different neural text generation tasks—such as summarization and machine translation—may present unique evaluation challenges and yield different conclusions. Second, the human evaluation component in our work was conducted by the authors themselves, potentially introducing expertise bias. Evaluators familiar with the field may interpret quality dimensions differently than end-users would. Finally, while our statistical methodology quantifies robustness against certain deviations from i.i.d. assumptions, real-world evaluation scenarios often involve more complex dependencies that require further methodological developments. Despite these limitations, we believe our work makes a substantial contribution to the field of text generation evaluation and provides a solid foundation for more statistically sound multi-criteria benchmarking approaches.

\section*{Ethics Statement}
This study uses only publicly available datasets that contain no personally identifiable information. Human evaluation was carried out by the authors on anonymized text continuations, ensuring that the underlying decoding strategies remained obscured. We acknowledge the potential ethical concerns associated with language models for text generation, particularly the risk of producing harmful content—whether through intentional misuse or unintended biases stemming from the training data and algorithms. We confirm that no conflicts of interest have influenced the outcomes, interpretations, or conclusions of this research. All funding sources are fully disclosed in the acknowledgments section. In the interest of transparency and reproducibility, we have thoroughly documented our methodology, experiments, and results and commit to sharing our code, data, and other relevant resources to further advance research in the field.

\section*{Acknowledgments}
Hannah Blocher received financial and general support via a scholarship from Evangelisches Studienwerk Villigst e.V. Julian Rodemann acknowledges support by the Federal Statistical Office of Germany within the co-operation project "Machine Learning in Official Statistics" as well as by the Bavarian Institute for Digital Transformation (bidt) and the Bavarian Academy of Sciences (BAS) within a graduate scholarship. Matthias Aßenmacher received funding from the Deutsche Forschungsgemeinschaft (DFG, German Research Foundation) as part of BERD@NFDI, under grant number 460037581.

\bibliography{colm2025_conference}
\bibliographystyle{colm2025_conference}

\clearpage

\appendix
\section{Appendix}

\subsection{Decoding strategies}
\label{a:decoding_strategies}

\begin{table}[H]
\centering
\begin{tabular}{p{2.5cm} l p{3.5cm} p{2cm}}
\hline
\textbf{Strategy} & \textbf{Parameters} & \textbf{Description} & \textbf{Authors} \\
\hline
Beam search & beam width = 5 & Deterministic search algorithm that maintains multiple hypotheses (beams). & \cite{Freitag_2017} \\
Contrastive search & $k = 10$, $\alpha = 0.6$ & Balances token probability and diversity through a contrastive objective. & \cite{su2022contrastive} \\
Sampling with temperature & temperature = 0.9 & Adjusts the sharpness of the probability distribution before sampling. & \cite{ackley1985learning} \\
Top-$k$ sampling & $k = 50$ & Samples from the $k$ most probable tokens. & \cite{fan2018hierarchical} \\
Top-$p$ sampling & $p = 0.95$ & Samples from the smallest set of tokens whose cumulative probability exceeds $p$. & \cite{holtzman2019curious} \\
\hline
\end{tabular}
\caption{Overview of evaluated decoding strategies and hyperparameter choices, following best performance reported by \cite{garces-arias-etal-2025-decoding}.}
\label{tab:decoding_strategies}
\end{table}

\subsection{Automatic metrics}\label{a: autom_metrics}

\paragraph{Diversity.} This metric aggregates $\mathrm{n}$-gram repetition rates: $$\texttt{div}=\prod_{n=2}^4 \frac{\mid \text { unique } \mathrm{n} \text {-grams }\left(\mathrm{x}_{\text {cont }}\right) \mid}{\mid\text { total } \mathrm{n} \text {-grams }\left(\mathrm{x}_{\text {cont }}\right) \mid}$$ A low diversity score suggests the model suffers from repetition, and a high diversity score means the model-generated text is lexically diverse.

\paragraph{Coherence.} Proposed by \citet{su2022contrastive}, the coherence metric is defined as the averaged log-likelihood of the generated text conditioned on the prompt as

$$
\texttt{coh}(\hat{\boldsymbol{x}}, \boldsymbol{x})=\frac{1}{|\hat{\boldsymbol{x}}|} \sum_{i=1}^{|\hat{\boldsymbol{x}}|} \log p_{\mathcal{M}}\left(\hat{\boldsymbol{x}}_i \mid\left[\boldsymbol{x}: \hat{\boldsymbol{x}}_{<i}\right]\right)
$$

where $\boldsymbol{x}$ and $\hat{\boldsymbol{x}}$ are the prompt and the generated text, respectively; [:] is the concatenation operation and $\mathcal{M}$ is the OPT model (2.7B) \cite{zhang2022optopenpretrainedtransformer}. 

\paragraph{Generation Perplexity.} 
The perplexity \( \texttt{ppl}(W) \) of a sequence of words (or tokens) \( W = w_1, w_2, ..., w_N \) is computed as \citep{10.1121/1.2016299,holtzman2019curious}:

\[
\texttt{ppl}(W) = \exp\left(-\frac{1}{N} \sum_{i=1}^{N} \log p(w_i \mid w_1, ..., w_{i-1}) \right)
\]

Here, \( p(w_i \mid w_1, ..., w_{i-1}) \) is the probability of word \( w_i \) given its preceding context. 

Perplexity measures how well a probabilistic model predicts a sequence of words. Lower perplexity indicates better predictive performance, as the model assigns a higher probability to the actual sequence. It is commonly used to evaluate the quality of language models.

\subsection{Q*Text}\label{a:qtext_technical}
Q*Text \citep{arias2025betteropenendedtextgeneration} is calculated based on normalized and penalized coherence, diversity, and generation perplexity (see Section \ref{a: autom_metrics}).

\paragraph{Metric Formulation} Q*Text is defined as:
\begin{equation}
\text{Q*Text} = \frac{\sum_{i=1}^{3} w_i M_i P_i(M_i)}{\sum_{i=1}^{3} w_i}
\end{equation}
where $M_i$ are normalized metrics, $w_i$ are weights, and $P_i(x) = \exp(-\alpha_i (x - \mu_i)^2)$ are Gaussian penalties that discourage extreme values. Parameters $\mu_i$ represent optimal targets while $\alpha_i$ controls penalty strength.

\paragraph{Normalization} We apply inverse normalization to perplexity (lower is better): $M_1 = \frac{p_{\max} - p_i}{p_{\max} - p_{\min}}$, and standard min-max normalization to coherence and diversity (higher is better): $M_j = \frac{m_j - m_{\min}}{m_{\max} - m_{\min}}$ for $j \in \{2,3\}$.

\paragraph{Parameter Optimization} The nine parameters $\theta = \{w_i, \mu_i, \alpha_i\}_{i=1}^{3}$ are optimized via:
\begin{equation}
\theta^* = \text{argmax}_{\theta} \rho_s(\text{Q*Text}(\theta), H)
\end{equation}
where $\rho_s$ is Spearman correlation and $H$ are publicly available human ratings \citep{garces-arias-etal-2025-decoding}.

\

\subsection{Human evaluation}
\label{a:human_evaluation}

\subsubsection{Instructions for human evaluators}
\label{a:defitinions}

Please disregard formatting characters and special characters such as <|endoftext|> or characters that have remained unrecognized and received unusual encoding. The evaluation should focus primarily on the quality of the content.

\begin{itemize}
    \item \textbf{Quality} should be measured by how human-like, fluent, and coherent the text is perceived by you.
    \item \textbf{Coherence:} The text feels consistent throughout, not a collection of jumbled topics. It maintains focus with a consistent thread and does not read as a series of disconnected sentences.
    \item \textbf{Fluency:} The text is written in grammatical English. There are no obvious grammar mistakes that a person would not typically make.
\end{itemize}

An incomplete final word or incomplete sentence should not be counted as a mistake and should not affect the fluency assessment. The English should be considered natural as long as it is grammatically correct. Do not penalize for spaces between parts of words (e.g., ``don 't'') or simpler sentences. Simple English is to be considered equally valid as complex English. Please utilize the following Likert scale.

\begin{table}[htbp]
\centering
\begin{tabular}{|c|l|p{0.6\textwidth}|}
\hline
\textbf{Score} & \textbf{Quality Level} & \textbf{Description} \\
\hline
5.0 & Excellent & Text is exceptionally clear, coherent, and well-structured. Content is comprehensive, accurate, and presented in a highly engaging manner. No improvements needed. \\
\hline
4.0 & Very Good & Text is clear, well-organized, and contains few errors. Ideas flow logically with appropriate transitions. Content is accurate and thorough. \\
\hline
3.0 & Good & Text communicates the intended message effectively. Organization is adequate with some minor clarity or coherence issues. Content is mostly accurate. \\
\hline
2.0 & Fair & Text has significant issues with clarity, organization, or accuracy that impact comprehension. Ideas may be underdeveloped or poorly connected. \\
\hline
1.0 & Very Poor & Text is difficult to understand with major structural problems, significant errors, and/or incomplete information. Communication largely fails. \\
\hline
\end{tabular}
\caption{Text quality assessment scale for human evaluators.}
\label{tab:quality-scale}
\end{table}

\subsubsection{Inter-rater agreements}
\label{a:inter_agreement}

The analysis focused on weighted agreement measures appropriate for ordinal data. The weighted Cohen's Kappa coefficient was 0.324, indicating fair agreement between evaluators when accounting for the magnitude of disagreements. This measure applies linear weights to disagreements based on their distance on the Likert scale, recognizing that a disagreement between ratings of 1 and 3 represents a larger discrepancy than between 1 and 2.

Spearman's rank correlation coefficient was 0.518, demonstrating a moderate positive correlation between the evaluators' ratings. This indicates that while absolute scores sometimes differed, the relative ranking of text quality showed reasonable consistency between evaluators. Additional analysis revealed that 82.7\% of all ratings were within one point of each other, with an average absolute difference of 0.82 points.

\begin{figure}[ht]
  \centering
  \includegraphics[width=0.60\textwidth]{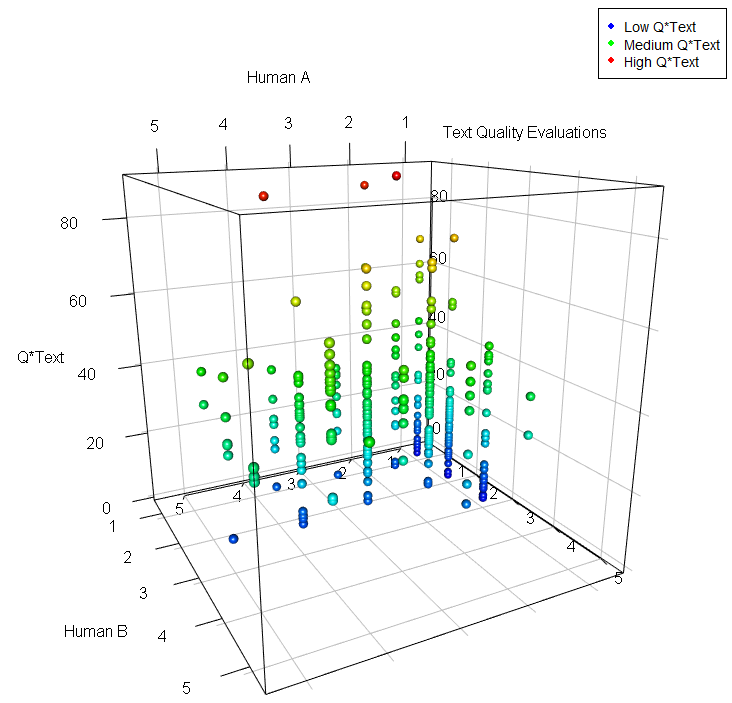}
\caption{Evaluation results for 300 text continuations generated from 50 prompts derived from Wikitext and Wikinews datasets. The assessment combines cardinal automatic metrics (Q*Text) with ordinal evaluations from two independent human raters using a 5-point Likert scale.}
\label{fig:evaluation}
\end{figure}

\end{document}